%% file: main.tex
\pdfoutput=1

\documentclass[11pt]{article}

\usepackage[]{acl}

\usepackage{times}
\usepackage{latexsym}

\usepackage[T1]{fontenc}

\usepackage[utf8]{inputenc}

\usepackage{microtype}
\usepackage{algorithm}
\usepackage{algpseudocode}

\usepackage{caption,subcaption}
\usepackage{amssymb}
\usepackage{graphicx}
\usepackage{amsmath}
\usepackage{multirow, epstopdf}
\usepackage{amssymb}
\title{RedHerring Attack: Testing the Reliability of Attack Detection}
\author{Jonathan Rusert
\\
  Purdue University, Fort Wayne\\
  \texttt{jrusert@pfw.edu }}
\date{July 2023}

\begin{document}

\maketitle

\input{abstract}

\input{introduction}

\input{threatmodel}
\input{attackdetection}
\input{method}

\input{experiments}

\input{humanstudy}
\input{llm}
\input{additionalmetricsMain}

\input{detectorattack}

\input{defense}

\input{adversarialtraining}
\input{conclusion}
\input{limitations}
\input{ethics}

\bibliography{main}
\bibliographystyle{acl_natbib}

\appendix
\input{relatedwork}
\input{motivatingexample2}
\input{uapadweights}
\input{datasetstatistics}
\input{originalResults}
\input{humanstudyinstructions}
\input{confidencecheck}
\input{defenseresults}
\input{examples}

\end{document}

%% file: abstract.tex
\begin{abstract}
    In response to adversarial text attacks, attack detection models have been proposed and shown to successfully identify text modified by adversaries. Attack detection models can be leveraged to provide an additional check for NLP models and give signals for human input. However, the reliability of these models has not yet been thoroughly explored. Thus, we propose and test a novel attack setting and attack, \textit{RedHerring}. \textit{RedHerring} aims to make attack detection models unreliable by modifying a text to cause the detection model to predict an attack, while keeping the classifier correct. This creates a tension between the classifier and detector. If a human sees that the detector is giving an ``incorrect'' prediction, but the classifier a correct one, then the human will see the detector as unreliable. We test this novel threat model on 4 datasets against 3 detectors defending 4 classifiers. We find that \textit{RedHerring} is able to drop detection accuracy between 20 - 71 points, while maintaining (or improving) classifier accuracy. As an initial defense, we propose a simple confidence check which requires no retraining of the classifier or detector and increases detection accuracy greatly. This novel threat model offers new insights into how adversaries may target detection models. 
\end{abstract}

%% file: introduction.tex
\section{Introduction}
\begin{figure}
    \centering
    \includegraphics[width=0.5\textwidth]{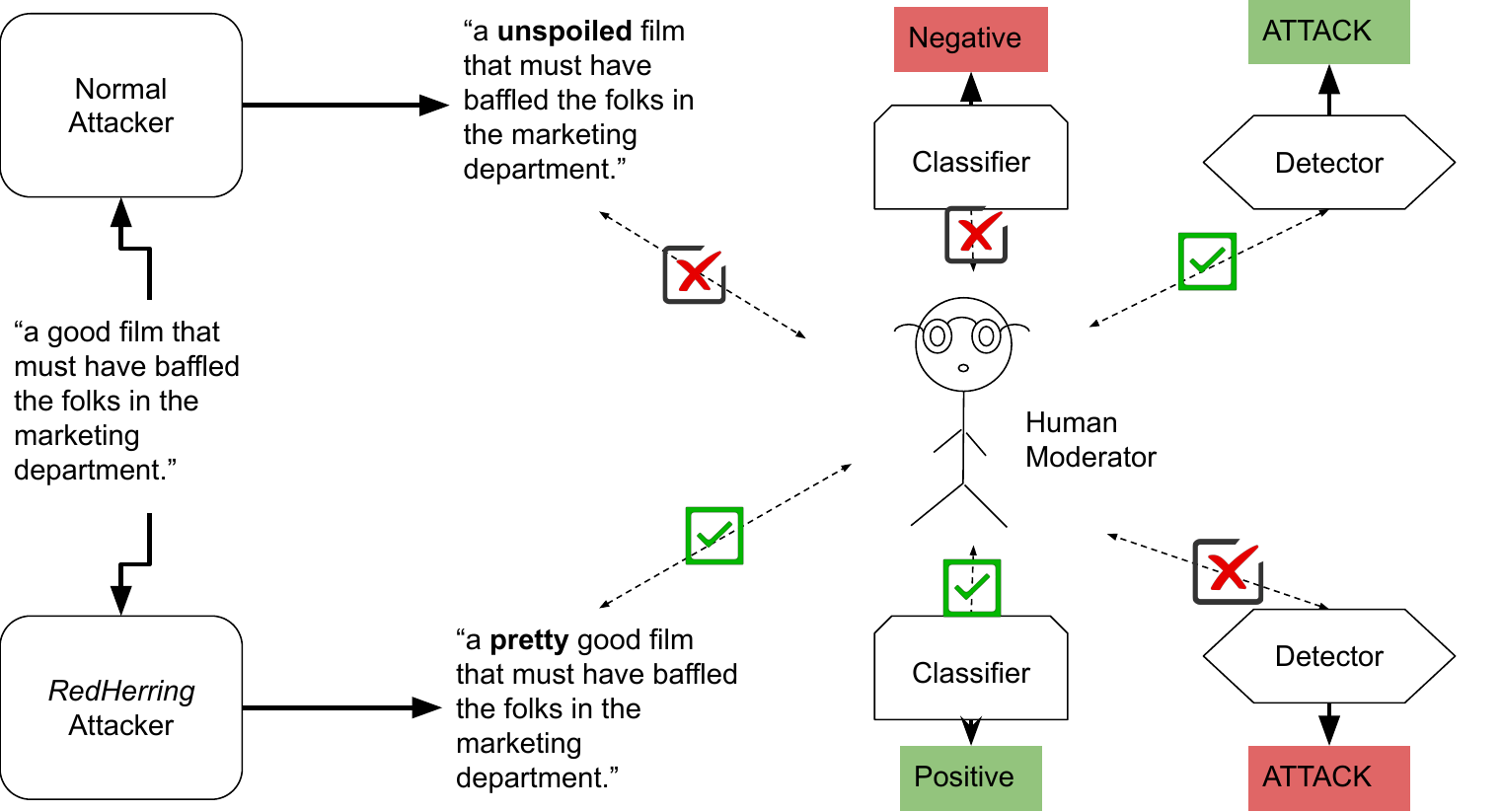}
    \caption{Visualization of \textit{RedHerring} attack. Normal adversarial text attacks focus on causing the text classifier to fail. \textit{RedHerring} aims to cause unreliability in the attack detector instead. This reliability happens since the detector indicates an attack, but the classifier is still classifying correctly, thereby creating doubt when a human examines both. }
    \label{fig:redherringattack}
\end{figure}
As use of social media and related platforms continues to grow, companies have increasingly leveraged automated text classifiers to help monitor content. These classifiers are essential for reducing the work load of human moderators. When classifiers  hinder users from posting content, users may try to subvert the classifier by modifying the text of their post. These ideas have inspired research in adversarial text attacks and defenses against attacks. 

Adversarial text attacks have been researched to measure multiple aspects of Natural Language Processing (NLP). These include classifier robustness \cite{hsieh-etal-2019-robustness} and concerns of censorship and privacy \cite{xie-hong-2022-differentially}. 
In response, defenses have been researched \cite{Li2021SearchingFA,li-etal-2023-text,wang-etal-2023-rmlm}. Defenses allow a classifier to correctly classify text even when it has been modified by an adversary. 
While defenses improved the robustness of classifiers, they often require retraining of models \cite{Zhu2020FreeLB,wang2020Infobert} or increase the number of queries required to make a classification \cite{Zeng2021CertifiedRT,rusert2022sample}. Thus, attack detection helps as an additional line of defense for classifiers. 

Attack detection algorithms aim to classify whether a text has been modified by an adversarial algorithm to subvert a classifier \cite{mozes-etal-2021-frequency,mosca-etal-2022-suspicious}. These algorithms  have been proposed to help point out potentially modified texts to human moderators. Thus, a website could leverage these algorithms to flag texts for humans to check against the automated classifier. A potential problem arises if the detection algorithm flags too many texts incorrectly, as this will cause human moderators' work load to greatly increase. 

In this research, we aim to investigate the vulnerability of text attack detection algorithms to attacks which cause distrust between humans and the algorithms. That is, we propose an attack, \textit{RedHerring}, which causes the attack detection system to trigger, without causing the classifier to fail. This type of attack is different from normal adversarial attacks in texts which only focus on failure of the NLP system. It is similar to other types of security attacks that cause unreliability in a system or algorithm \cite{qi-2021-ddos,xu-2019-dos}. 

Our research makes the following contributions:

1. Introduce a novel attack \textit{RedHerring} and novel text attack setting which measures the vulnerability of attack detection algorithms to unreliability attacks.

2. Test \textit{RedHerring} on 4 datasets against 3 detectors defending 4 classifiers. We find that \textit{RedHerring} is able to drop detection accuracy for 39 points against WDR on average, 61 points against FGWS, and 36 points against UAPAD. Furthermore, \textit{RedHerring} maintains or increase classification accuracy which pits the classifiers and detectors against each other. 

3. Verify \textit{RedHerring} with human annotations and experiments which highlight how it differs from previous attack research.

4. Propose and test an initial defense against this type of attack which leverages the confidence score of the classifier and requires no retraining of models. Though it does not fully restore detection accuracy, we find this defense greatly mitigates the attack, restoring up to 80\% detection accuracy.

The strength of \textit{RedHerring} attack demonstrates the need for greater robustness of attack detection algorithms. We share the code and examples for future research purposes (\url{https://github.com/JonRusert/RedHerringAttack}).

%% file: threatmodel.tex
\section{Proposed Threat Model}
Here we define the problem and then the novel threat model explored with \textit{RedHerring} attack. 

\textbf{Text Classifiers and Attack Detectors:} Let $X$ represent a text, $f()$ represent a classifier, and $g()$ represent a detector.

The goals of text classifiers are to assign a label, from a set of labels $Y = \{y_0, y_1, ..., y_n\}$ to a given text, $f(X_i) = y_i$. Classifiers are trained on a set of texts and assigned labels, generally leveraging embedding representations of the texts. Adversarial attacks target text classifiers by making slight perturbations to an input text, such that it causes the classifier to fail, while at the same time aiming to maintain original meaning to human readers. This is often done by replacing, adding, or rewriting words in the input text, that the text classifier is most reliant on.

The goals of attack detectors are to label an input text as being modified by an adversarial attack or not, $g(X_i) = \{ATTACK, NOT\}$. Unlike text classifiers, attack detectors do not necessarily train new models to accomplish this task nor do they train on the text input itself, instead they often leverage the knowledge of the classifier to label the input text.

\textbf{Threat Model:} We assume that the detector and classifier are deployed to automatically monitor a website's posts. The text may be instantly rejected if it does not pass the classifier check. Texts that pass the classifier are double checked by the detector. If the detector flags the text as adversarial, a human moderator checks the text and the classifier/detector's labels. Note that this model is set up to reduce queries as detectors often leverage the classifier as well (sometimes multiple times), to make their decision. 

We assume that \textit{RedHerring} (Section \ref{sect:rhattack}) has black-box probability access to the classifier and detector. That is, it is able to send queries to both models and receive feedback in terms of labels and confidence scores. This black-box assumption follows similarly to previous adversarial attack research \cite{alzantot2018generating, garg2020bae, li-etal-2020-bert-attack,gao2018blackbox, hsieh-etal-2019-robustness, li2021contextualized}.

\section{Motivation for Studying Attack}
The goal of \textit{RedHerring} is to cause the detector to become unreliable in the eyes of the company or human moderator. This could cause financial issues (as the company may spend additional funds to find a new detector) or vulnerability issues (the company may abandon attack detectors entirely). \textit{RedHerring} aims to accomplish this by causing the detector to make ``false flags'', while the classifier continues to work correctly (outlined in Section \ref{sect:attackgoals}). That is, the detector will predict an attack is happening when there does not appear to be one (classifier is classifying correctly). This can cause the human moderator to distrust the detection algorithm. 
Note that this threat model is different than simply causing the detector (and classifier) to fail, which is more similar to original text adversarial attacks. 

\textbf{Motivating Example 1:} Consider the case of social media websites that aim to protect against hate speech. The classifier would try to flag hate speech. A normal adversary would try to still write their hate speech by targeting the classifier during their attack. An attack detector would help protect against potential attacks by flagging them so that human moderators could double-check them. However, \textit{RedHerring} could be deployed to cause uncertainty with the attack detector. By causing the attack detector to send flags to the human moderators who check and see that the post is hate speech and the classifier is classifying it as hate speech, the moderators would lose trust and also time in additional moderation.

\textbf{Motivating Example 2:} Another motivating example in the form of spam detection. Due to space, this is expanded on in the appendix (Appendix \ref{sect:motivatingexample}).

%% file: attackdetection.tex
\section{Attack Detection}\label{sect:attackdetection}
To evaluate \textit{RedHerring} we examine three attack detection algorithms\footnote{Full Related Work found in Appendix \ref{sect:relatedwork}}: WDR \cite{mosca-etal-2022-suspicious}, FGWS \cite{mozes-etal-2021-frequency}, and UAPAD \cite{gao-etal-2023-universal}. 
These algorithms were chosen because of their strong performance and variation of methodology.

1. Word-level Differential
Reaction (WDR) - WDR detects attacks by examining the change in classifier logits when a word is removed from the input text. When adversarial words are removed, the change is more likely to be negative. This change is referred to as the Word-level Differential Reaction (WDR). A classifier is trained to detect an adversarial attack based on the sorted WDRs for each text. The authors find XGBoost to perform best, thus we use this as the classifier for WDR. 

2. Frequency-Guided Word Substitutions (FGWS) - FGWS detects attacks by replacing infrequent words in input text with frequent synonyms. The frequency is determined by the training data. The synonyms are obtained via WordNet \cite{miller-1994-wordnet} and Glove Word Embeddings \cite{pennington2014glove}. For every word below a chosen log frequency threshold, $\delta$, in an input text ($X$), FGWS replaces the word with a more frequent synonym. This modified text, $X'$, is then compared to $X$ via the classifiers ($f()$) prediction scores on the originally predicted class $y$. If $f(X)_y - f(X')_y > \gamma$, then the text is considered an adversarial attack. Note $\gamma$ is another chosen hyperparameter. Following the authors, and confirming via preliminary experiments, we choose $\delta = 0.9$ and $\gamma = 0.05$. 

3. Universal Adversarial Perturbation Attack Detection (UAPAD) - UAPAD detects attacks by utilizing universal adversarial perturbations (UAPs). UAPs are single perturbations which can cause models to fail when added to the inputs. UAPAD recognizes that adversarial samples are often closer to the decision boundary than clean examples. Thus, they add UAPs to input texts to help determine clean versus adversarial samples. Specifically, UAPAD makes 2 calls to the classifier, one call on the original text and one on the UAP text. If the labels are not equal, it is labels as an attack. Note that each classifier/dataset requires a different UAP delta weight. To test our attack in the strongest setting, we choose the weights which result in the strongest original detection accuracy for UAPAD (Found in Appendix \ref{sect:uapadweights}).

\subsection{Detection Experiments and Results} \label{sect:dectExperiments}
To verify the effectiveness of the above attack detection algorithms (or detectors), we evaluate the detectors on attacks across 4 datasets. A subset of these datasets (the class distributions can be found in Table \ref{tab:datasetstatistics}, in the Appendix) were evaluated in each of the original papers:
1. AG News - Classification on news text on one of four categories: Sports, Business, World, Sci/Tech.  
2. IMDB - Binary sentiment classification on movie reviews from IMDB website.
3. Rotten Tomatoes (RT) - Binary sentiment classification on movie reviews from Rotten Tomatoes.
4. SST-2 - Stanford Sentiment Treebank, contains binary sentiment classification on movie reviews. 

We evaluate the detectors against attacks carried out by PWWS \cite{ren-etal-2019-generating}. PWWS modifies text by selecting WordNet synonym replacements which cause the largest prediction probability drops. PWWS also leverages word saliency to help rank appropriate replacements. Note that we focus on this attack as FGWS is attacker agnostic, WDR demonstrates the transferability of a detector trained on one attack against others in their own research, and UAPAD tests against PWWS.

We focus on 3-4 classifiers for each dataset, obtained mainly from the TextAttack \cite{morris2020textattack} library: Albert \cite{lan2019albert}, RoBERTa \cite{liu2019roberta}, DistilBERT\cite{sanh2020distilbert}, BERT \cite{devlin-etal-2019-bert}, and T5 \cite{raffel2020t5}\footnote{Due to incompatability with UAPAD (see Section \ref{sect:llm}), UAPAD was excluded for T5.}. 

We measure the accuracy obtained for each detector on each dataset. For AG News, IMDB, and RT we test on 1000 adversarially modified texts and 1000 original texts. For SST-2 we test on 436 modified and 436 non-modified due to the limited size of the test set. 

For FGWS, frequencies are learned from the training datasets and synonyms are obtained from WordNet and from Glove\footnote{https://nlp.stanford.edu/projects/glove/}. For WDR, we train a detector on the WDRs of each specific dataset. For UAPAD, we first obtain BERT UAPs to verify correct usage of provided code, then follow the same process for the other classifiers. We also choose the strongest UAP delta weights (Appendix \ref{sect:uapadweights}). Thus, purposely giving the detectors a stronger advantage against our proposed attack.

\begin{table}[]
    \centering
    \footnotesize
    \begin{tabular}{c|c|c||c||c}
         & Class. & WDR & FGWS & UAPAD\\\hline
         \parbox[t]{2mm}{\multirow{3}{*}{\rotatebox[origin=c]{90}{AG}}} & Albert & 94.0 & 89.8 & 94.1\\
          & RoBERTa & 92.3 & 88.7 & 90.8\\
         & DistilBERT & 93.3 & 91.5 & 94.1\\\hline

         \parbox[t]{2mm}{\multirow{3}{*}{\rotatebox[origin=c]{90}{RT}}} & Albert & 55.5 & 81.8 & 81.1\\
         & RoBERTa & 77.1 & 83.7 & 79.2 \\
         & BERT & 60.2 & 82.6 & 78.8\\
         & T5 & 75.6 & 82.2 & -\\\hline

         \parbox[t]{2mm}{\multirow{3}{*}{\rotatebox[origin=c]{90}{IMDB}}} & Albert & 84.9 & 85.4 & 70.0\\
          & RoBERTa & 88.2 & 90.5 & 77.0\\
         & DistilBERT & 80.3 & 82.2 & 73.8\\\hline

         \parbox[t]{2mm}{\multirow{3}{*}{\rotatebox[origin=c]{90}{SST-2}}} & Albert & 72.5 & 82.3 & 81.9\\
         & RoBERTa & 80.8 & 82.1 & 83.1\\
         & DistilBERT & 64.0 & 63.0 & 69.4\\
         & T5 & 80.4 & 81.8 & - \\\hline
         
    \end{tabular}
    \caption{Attack Detection results. Accuracy shown. WDR uses XGBoost as detector. }
    \label{tab:detectionResults}
\end{table}

\noindent \textbf{Detection Results:} The results for all detectors can be found in Table \ref{tab:detectionResults}. As can be seen, FGWS obtains 80 - ~90\% accuracy scores across datasets, which is consistent (or even stronger) with reported results. WDR performs well on AG News (as reported) and varies on the other measured datasets, and UAPAD performs similarly to reported results\footnote{UAPAD only used BERT, so it is difficult for direct comparisons.} Note that weaker scores on RT are consistent with original reported scores. This confirms the effectiveness of all detection algorithms. 

%% file: method.tex
\section{\textit{RedHerring} Attack}\label{sect:rhattack}

Let $X = \{t_1, t_2, ..., t_n\}$ be a text of length $n$. Additionally, let f(X) represent the text classifier which assigns a class label $l = \{c_1, ..., c_k\}$ to a given $X$.  Also, let g(X) represent an adversarial attack detector which assigns a detection label $d = \{NOT, ATTACK\}$ to the given $X$. Finally, let rh(X) represent our proposed attack which takes in $X$ and produces $X'$, a modified form of $X$.

\begin{table*}[]
    \centering
    \footnotesize
    \begin{tabular}{c|c||c|c|c|c|c|c|}
          & & \multicolumn{6}{c}{Perturbed Accuracy} \\\hline
          
          & Classifier & \multicolumn{3}{c|}{Classifier Acc.} & \multicolumn{3}{|c}{Detection Acc.}\\\hline
         
         \multicolumn{2}{c|}{RH Target:}   & WDR ($\Delta$) & FGWS ($\Delta$) & UAPAD($\Delta$) & WDR ($\Delta$) & FGWS ($\Delta$) & UAPAD ($\Delta$)\\\hline

         
         \parbox[t]{2mm}{\multirow{3}{*}{\rotatebox[origin=c]{90}{AG}}} & Albert &  96.8 (-2.6) & 97.6 (-3.4) & 97.7 (-3.5) & 60.9 (25.1) & 31.5 (50.4) & 42.7 (47.7)\\
         & RoBERTa & 96.6 (-1.9) & 97.0 (-2.3) & 96.4 (-1.7) & 55.6 (29.9) & 39.2 (40.8) & 42.8 (42.5)\\
         & Distilbert &  96.1 (-1.7)	 & 97.0 (-2.6) & 97.0 (-2.6) & 65.1 (24.2) & 34.0 (54.1) & 40.6 (45.6)\\\hline
         \multicolumn{2}{c||}{Average $\Delta$} & -2.1 & -2.8 & -2.6 & 26.4 & 48.4 & 45.3\\\hline\hline
        
         \parbox[t]{2mm}{\multirow{3}{*}{\rotatebox[origin=c]{90}{IMDB}}}& Albert & 95.9 (-2.4) & 97.6 (-4.1) & 98.6 (-5.1) & 62.1 (18.3)& 23.3 (65.7) & 54.4 (19.0)\\
         & RoBERTa & 96.5 (-1.4)	& 98.9 (-3.8) & 98.4 (-3.3) & 75.4 (12.1) & 21.6 (70.5) & 65.2 (22.2)\\
         & Distilbert & 93.8 (-1.8)  & 98.9  (-6.9)  & 98.3 (-6.3) & 60.0 (30.6)  & 9.1 (77.4) & 49.3  (20.2)\\\hline
         \multicolumn{2}{c||}{Average $\Delta$} & -1.9 & -4.9 & -4.9 & 20.3 & 71.2 & 20.5\\\hline\hline

         \parbox[t]{2mm}{\multirow{3}{*}{\rotatebox[origin=c]{90}{RT}}}& Albert & 93.0 (-8.2)  & 95.7 (-10.9)  & 95.7 (-10.9)  & 6.0 (48.8)  & 6.5 (69.9) & 17.2 (51.5)\\
         & RoBERTa & 95.6 (-7.0) & 97.5 (-8.9) & 97.5 (-8.9)  & 15.2 (61.5)  & 14.1 (76.7)  & 37 (46.2)\\
         & BERT & 94.0 (-8.6) & 96.1 (-10.7) & 96.1 (-10.7) & 16.3 (50.2) & 20.9 (63.6)& 24.2 (49.1) \\\hline
          \multicolumn{2}{c||}{Average $\Delta$} & -7.9 & -10.2 & -10.2 & 53.5 & 70.1 & 48.9\\\hline\hline

    
         \parbox[t]{2mm}{\multirow{3}{*}{\rotatebox[origin=c]{90}{SST-2}}}& Albert &  92.7 (0.0) & 98.2 (-5.5) & 98.4 (-5.7)  & 8.4 (48.9)  & 10.0 (73.7)  & 26.0 (31.0) \\
          & RoBERTa & 94.0 (0.0)  & 98.9 (-4.9)  & 98.9 (-4.9)  & 4.8 (69.5)  & 31.5 (59.6)  & 33.3 (50.9)\\
         & Distilbert & 55.7 (0.0) & 84.9 (-29.2)  & 84.9 (-29.2)  & 12.0 (45.0)  & 3.0 (45.7) & 18.3 (42.1)\\\hline
         \multicolumn{2}{c||}{Average $\Delta$} & 0.0 & -13.2 & -13.3 & 54.5 & 59.7 & 41.3\\\hline\hline

        \multicolumn{2}{c||}{Overall Avg. $\Delta$} & -3.0 & -7.8 & -7.7 & 38.7 & 62.3 & 39.0\\\hline

    \end{tabular}
    \caption{\textit{RedHerring} Attack Results. RH Target - indicates the target of \textit{RedHerring} Attack. $\Delta$ indicates the difference between the original accuracy and the perturbed accuracy. Average $\Delta$ indicate the average difference between the original and perturbed accuracy values. Overall Avg. $\Delta$ indicates the average drops including all combinations in that column (across datasets and classifiers).}
    \label{tab:redherringResults}
\end{table*}

\subsection{Attack Goal}\label{sect:attackgoals}
The overall goal of our proposed attack is to cause unreliability of the attack detection algorithm. Hence, it is designed to maintain 3 subgoals: 

\begin{enumerate}
    \item $g(X') == ATTACK$
    \item $f(X) == y_{true}$ 
    \item X' maintains original meaning to humans.
\end{enumerate}

With the first subgoal, the attack aims to cause the attack detection algorithm to trigger and note to the user an attack is occurring. With the second, the attack aims to keep the original classifier classifying the correct label. Finally, the third subgoal requires humans to understand X' the same as X.

All three subgoals are needed for our attack to work successfully. If the first subgoal is met, but the second is not (the classifier misclassifies), then the detection algorithm is working as intended. If the first and second are met, but not the third (the output changes meaning to humans), then the human may view the classifier as unreliable rather than the detector, or may conclude that the detector is acting correctly. 

The third goal naturally leans towards word level or sentence level modifications, as character-level will appear more obviously to the human that the text is under attack. Hence,all these goals are considered in the overall design of the attack.

\subsection{Attack Design}
Our attack follows similarly to previous word-based attacks and can be divided into two steps: selection and addition. 

\subsubsection{Selection}
For selection, we adapt the greedy selection method (also sometimes named Word Importance Score) from previous text attack research \cite{hsieh-etal-2019-robustness}. Greedy select removes one word (or token) at a time and notes the classifier's change in probability:

\begin{equation}
    GS = f(X) - f(X//w_m)
\end{equation}

The change in probabilities are then sorted and the position that caused the greatest drop in probability is chosen for replacement. Note that even though our end goal is to attack the detector, we still leverage the classifier to choose the word. This is because most attack detection systems examine classifiers' probability changes themselves. Furthermore, many of the detectors perform multiple queries to the classifier during detection. Thus, if we were to perform a similar selection above to the detectors, then the queries would increase exponentially. 




\subsubsection{Addition}
Similarly to selection, we build of similar previously leveraged methods. Specifically, we build off attacks such as BAE \cite{garg2020bae} and BERT-ATTACK \cite{li-etal-2020-bert-attack} which leverage a BERT based model to infill masked tokens. This is feasible since BERT is trained on is Masked Language Modelling, where input tokens are randomly masked and BERT predicts what the masked token should be. Hence, we can simulate this by adding a mask token next to the selected word from the previous step and asking the BERT model for word suggestions. Instead of pure replacement, we add a mask token next to the word to maintain semantic integrity while increasing the chance for the detector to be triggered.

Recall that we still want to maintain the previous subgoals, so we focus on the change in detector and check against the classifier:

\begin{equation}
    \begin{aligned}
        X_{repl} = \{{w_{1}},...,w_{repl},...,{w_{n}}\}\\
        \Delta g(X)_{repl} = g(X) - g(X_{repl}) \\
    \end{aligned}
\end{equation}
\begin{equation}
   f(X)_{repl} = f(X_{repl})
\end{equation}

We want the classifier ($f(X)_{repl}$) to remain correct, but the detector to become incorrect. Beginning with the top suggestion from BERT, we check the labels for each. If the suggestion ($w_{repl}$) causes the detector's label to flip, but keeps the classifier's label the same, it is chosen as the suggestion. Otherwise, the probability changes are noted and the next suggestion is checked. We check the top $m$ suggestions until either a suggestion is found which satisfies the subgoals or all $m$ have been checked. In the latter case, the suggestion which maximizes $\Delta g(X)_{repl}$ is added to the text and the process continues with the next selected word.

%% file: experiments.tex
\section{Experimental Results and Discussion}
We test \textit{RedHerring} attack on the same 4 datasets (AG News, IMDB, RT, SST-2) described in Section \ref{sect:dectExperiments}. Following previous attack research \cite{jin2020bert}, we sample up to 1000 examples for each dataset for modification. We run the attack\footnote{Attacks were run on 80-core 384G CPUs for a max time limit of 1 hour for efficiency and practicality purposes.} against all detection algorithms on each of the 3 noted classifiers for each dataset (Section \ref{sect:dectExperiments}). Note that since \textit{RedHerring} focuses on causing distrust with detectors, \textbf{stronger, newer classifiers would only cause the first goal of the attack to be easier to achieve} (This is confirmed by testing the attack against an LLM (Section \ref{sect:llm})).  We use the same detection hyperparameters chosen in Section \ref{sect:attackdetection}. Note that UAPAD does not return an actual detection score which makes it more difficult to attack. We approximate the score by focusing on the change in a class when the UAP is applied.

\textbf{Attack Metrics:} We measure accuracy of detectors and classifiers, change in accuracy caused by \textit{RedHerring}, the False Positive Rate of the detectors, and the cases where both the classifier was correct and detector incorrect (fooled by \textit{RedHerring}).

\subsection{\textit{RedHerring} Results} Table \ref{tab:redherringResults} contains the main results. The "Original Accuracy" results can be found in Table \ref{tab:originalResults} (Appendix \ref{sect:originalresults}). These denote the classification and detection scores against non-modified texts, while "Perturbed Accuracy" indicate the score of the same models on the text modified by \textit{RedHerring}. Note that the original detection scores differ from those in Section \ref{sect:dectExperiments} since the original scores in Table \ref{tab:redherringResults} are purely on non-modified texts. We make the following observations:

\noindent \textbf{\textit{RedHerring} causes drops in attack detection across datasets.}  For all detection algorithms, we see drops in detection accuracy across all datasets. For AG News, we see average drops of 26 against WDR, 48 against FGWS, and 35 against UAPAD. These results are similar for IMDB, average of 20 against WDR, 71 against FGWS, and 21 against UAPAD. Finally, for both RT and SST-2 datasets, we see effectiveness against WDR increase to drops of 54 and 55 for RT and SST-2 respectively. FGWS also sees strong drops for these datasets, 65 and 60 for RT and SST-2 respectively. UAPAD also increases in drops with 49 for RT and 41 for SST-2. These drops show the strength of \textit{RedHerring} in causing attack detection to trigger falsely. 

\noindent \textbf{\textit{RedHerring} increases classification accuracy across datasets.} A drop in detection accuracy is useless if the classifiers are dropping in accuracy as well, since this would mean that the detectors are predicting correctly. Thus, our second goal is to maintain (or improve) classification accuracy. When examining the drops of the classifiers\footnote{Negative drops indicate an increase in accuracy.}, we see improvements across datasets. We see average increases in accuracy of 3 against WDR, 8 for FGWS and UAPAD. This increase is most likely due to RedHerring's main goal of causing the detector to appear unreliable. Since RedHerring does not want the classifier to fail (as it targets the detector), when it chooses words to modify/replace, it rejects the words that would cause the classifier to fail the classification. Additionally, it does not add any constraints on words that would increase the classification probability.
These results satisfy the second goal of \textit{RedHerring} attack.

\noindent \textbf{FGWS most vulnerable to attack, WDR, UAPAD most resilient.}
When examining the overall drops across datasets, we see that \textit{RedHerring} is able to drop FGWS by 62 points on average, compared to 39 points for WDR and UAPAD. Part of this may be explained by WDR's examination of every word in the input text and UAPAD's acknowledgement that attacked texts are often close to the decision boundary. On the contrary, FGWS only replaces less frequent words. Thus, if \textit{RedHerring} finds words which are more frequent or those without synonyms, then FGWS will not replace these words in checking the text.

\noindent \textbf{\textit{RedHerring} attack is more effective against shorter texts.} The average lengths of the texts are 43, 215, 21, and 20 for AG News, IMDB, RT, and SST-2. We find that \textit{RedHerring} is the most effective against the shorter texts, causing an average drop of 56 for the RT detectors and 52 for SST-2. These averages are much higher than drops of AG News (37) and IMDB (37). This indicates greater vulnerability for detectors on shorter texts. This vulnerability is most likely caused by 2 reasons. First, since \textit{RedHerring} adds words, this is more likely to trigger the detection algorithms since a new word on a smaller text is a bigger change than for a larger text. Second, \textit{RedHerring} uses greedy select to determine where to add words in the text. For shorter texts, removing words will cause a clearer drop than with larger texts. Both of these make shorter texts more vulnerable.

%% file: humanstudy.tex
\subsection{Qualitative Analysis}

\subsubsection{Human Study}
To further gauge the effectiveness of \textit{RedHerring}, we employ 3 human annotators to label texts. The annotators are given the text, the classifier label, and the detector label. They are asked to determine if an attack is occuring or not (full instructions including verification of attack understanding in Appendix \ref{sect:humanstudyinstructions}). The annotators label the same 40 \textit{RedHerring} texts, 40 texts attacked by PWWS, and 40 non attacked texts. The majority vote is taken for each text. 

We find that \textit{RedHerring} is indeed able to create a larger disagreement and lack of trust with the detector. Specifically, human annotators agreed with the detector only 41\% of the time for \textit{RedHerring} texts, compared to 67\% of the time for PWWS texts, and 87\% for non attacked texts. The annotators had a fair agreement with each other achieving an average Cohen's Kappa score of 0.31. These results further highlight \textit{RedHerring}'s ability to cause distrust between the human and the detector. 

\subsubsection{BERTScore}
In addition to the human evaluation, we measure the semantic integrity of the generated text via BERTScore \cite{zhang2019bertscore}. The full results can be found in Table \ref{tab:bertscores}. As can be observed, for AG News, Rotten Tomatoes, and IMDB, the BERTScores are high ranging from 90 - 98, indicating strong similarity of the produced texts with the original. SST2 produced texts are lower around 80, most likely because modifications have a higher impact due to the short length of the texts. 

\begin{table}[]
    \centering
    \footnotesize
    \begin{tabular}{c|c|c|c}
         RH Target: & WDR & FGWS & UAPAD\\\hline
         \multicolumn{4}{c}{AG News}  \\\hline
         Albert & 97.9 & 91.4 & 90.3 \\
         Roberta & 97.6 & 90.7 & 90.9 \\
         Distilbert & 97.8 & 90.2 & 90.1 \\\hline
         \multicolumn{4}{c}{Rotten Tomatoes} \\\hline
         Albert & 96.8 & 95.6 & 95\\
         Roberta & 96.3 & 94 & 93.5\\
         BERT & 96.6 & 93.9 & 94.4\\
         T5 & 95.7 & 94.3 & -\\\hline
         \multicolumn{4}{c}{IMDB} \\\hline
         Albert & 97.1 & 95.6 & 95.2\\
         Roberta & 97.1 & 95.1 & 94.3\\
         Distilbert & 96.9& 94.6 & 94.4\\\hline
         \multicolumn{4}{c}{SST2} \\\hline
         Albert & 79.9 & 79.7 & 79.7\\
         Roberta & 80 & 79.5 & 79.6\\
         Dilstilbert & 80 & 80 & 79.9\\
         T5 & 79.8 & 79.7 & -\\\hline
    \end{tabular}
    \caption{BERTScores for the texts generated by RedHerring Attack for the examined datasets. }
    \label{tab:bertscores}
\end{table}

%% file: llm.tex
\begin{table}[]
    \footnotesize
    \centering
    \begin{tabular}{c|c||c|c|c|c}
                    
          & & \multicolumn{2}{c|}{T5 Acc.} & \multicolumn{2}{|c}{Detection Acc.}\\\hline
         
         \multicolumn{2}{c|}{RH Target:}   & WDR & FGWS & WDR & FGWS\\\hline

         \parbox[t]{2mm}{\multirow{2}{*}{\rotatebox[origin=c]{90}{RT}}}& Orig. & \multicolumn{2}{c|}{88.5} & 73.3 & 85.3\\
         & RedH. & 97.2 & 97.7 & 5.4 & 12.5 \\\hline
           \multicolumn{2}{c||}{Drops} & -8.7 & -9.2 & 67.9 & 72.8 \\\hline\hline

    
         \parbox[t]{2mm}{\multirow{2}{*}{\rotatebox[origin=c]{90}{SST2}}}&  Orig.  &  \multicolumn{2}{c|}{93.2}  & 72.8 & 87.4 \\
          & RedH. & 99.1 & 99.2 & 3.9 & 7.7\\\hline
         \multicolumn{2}{c||}{Drops} & -5.9 & -6.0 & 68.9 & 79.7\\\hline\hline

        \multicolumn{2}{c||}{Averages} & -7.3 & -7.6 &  68.4 & 76.3\\\hline

    \end{tabular}
    \caption{\textit{RedHerring} Attack Results against the T5 model. Orig. indicates the accuracy with no attack. RedH. indicates the accuracy after RedHerring attack.  Averages indicate average drops in that column.}
    \label{tab:t5Results}
\end{table}

\subsection{RedHerring Against LLM}\label{sect:llm}

We also verify RedHerring's strength by testing it against the T5 generative model \cite{raffel2020t5}. T5 is fine-tuned on the respective datasets, RT and SST-2, the label is generated by the model from the input text. For example, in SST-2, it generates ``Positive'' or ``Negative''. To obtain the logits needed for the detectors, we obtain the logits for the class tokens at that specific positive, and then pass just those logits  to the softmax function. UAPAD requires backpropagation after passing the logits through the softmax function, so it was found incompatible with the generative model in its current state, hence we focus on FGWS and WDR for these experiments. As previously seen, if the attack is effective for FGWS/WDR, it will also be effective against UAPAD. Results can be found in Table \ref{tab:t5Results}.

As seen, \textit{RedHerring} is indeed effective against the detectors which accompany T5. We see large drops for WDR and FGWS of 68+ points. Additionally, T5 does not lose its prediction ability, increasing at least 7 points in all instances.This further highlights the effectiveness of \textit{RedHerring}.

%% file: additionalmetricsMain.tex
\subsection{FPR and Number of Successes}
\begin{table}[]
    \centering
    \footnotesize
    \begin{tabular}{c|c|c|c|c}
 
        & Classifier & \multicolumn{3}{|c}{False Positive Rate} \\\hline
         
         \multicolumn{2}{c|}{RH Target:}    & WDR & FGWS & UAPAD \\\hline

         \parbox[t]{2mm}{\multirow{3}{*}{\rotatebox[origin=c]{90}{AG}}} & Albert & 	25.5 & 50.4 & 47.7\\
         & RoBERTa & 29.9 & 40.8 & 42.6\\
         & Distilbert & 24.2 & 54.1 & 45.6\\\hline\hline
        
         \parbox[t]{2mm}{\multirow{3}{*}{\rotatebox[origin=c]{90}{IMDB}}}& Albert  & 18.3 & 65.7 & 19\\
         & RoBERTa & 12.1 & 70.5 & 22.2\\
         & Distilbert & 30.6 & 77.4 & 20.2\\\hline\hline

         \parbox[t]{2mm}{\multirow{4}{*}{\rotatebox[origin=c]{90}{RT}}}& Albert & 48.8 & 55.5 & 51.5\\
         & RoBERTa & 61.5 & 76.7 & 46.2\\
         & BERT & 50.2 & 63.6 & 49.1\\
        & T5 & 67.9 & 72.8 & - \\\hline\hline

         \parbox[t]{2mm}{\multirow{4}{*}{\rotatebox[origin=c]{90}{SST-2}}}& Albert & 48.9 & 73.7 & 31.0 \\
          & RoBERTa & 69.5 & 59.6 & 50.9\\
         & Distilbert & 45 & 45.7 & 42.1 \\
         & T5 & 68.9 & 79.7 & - \\\hline\hline

    \end{tabular}
    \caption{ False positive rates for the detectors on the original test texts which have been perturbed by RedHerring. RH Target - indicates the target of \textit{RedHerring} Attack}
    \label{tab:FPRResults}
\end{table}

\begin{table}[]
    \centering
    \footnotesize
    \begin{tabular}{c|c|c|c|c}
 
        & Classifier & \multicolumn{3}{|c}{Overlapping Success} \\\hline
         
         \multicolumn{2}{c|}{RH Target:} & WDR & FGWS & UAPAD \\\hline

         \parbox[t]{2mm}{\multirow{3}{*}{\rotatebox[origin=c]{90}{AG}}} & Albert & 374 & 673 & 525	\\
         & RoBERTa & 429 & 594 & 556\\
         & Distilbert & 327 & 651 & 585\\\hline\hline
        
         \parbox[t]{2mm}{\multirow{3}{*}{\rotatebox[origin=c]{90}{IMDB}}}& Albert  & 294 & 628 & 478\\
         & RoBERTa & 193 & 651 & 455\\
         & Distilbert & 308 & 738 & 604\\\hline\hline

         \parbox[t]{2mm}{\multirow{4}{*}{\rotatebox[origin=c]{90}{RT}}}& Albert & 899 & 921 & 812 \\
         & RoBERTa & 837 & 856 & 627\\
         & BERT & 810 & 787 & 749\\
        & T5 & 938 & 873 & - \\\hline\hline

         \parbox[t]{2mm}{\multirow{4}{*}{\rotatebox[origin=c]{90}{SST-2}}}& Albert & 788 & 781 & 627\\
          & RoBERTa & 823 & 596 & 579\\
         & Distilbert & 718 & 729 & 703 \\
         & T5 & 835 & 803 & - \\\hline\hline

    \end{tabular}
    \caption{ Number of test examples where the classifier is correct, but the detector is incorrect (ie. RedHerring accomplished it goal).  RH Target - indicates the target of \textit{RedHerring} Attack}
    \label{tab:overlapResults}
\end{table}

We further quantify the success of \textit{RedHerring} by examining the False Positive Rates caused by \textit{RedHerring} and the raw number of ``successful'' attacks. These values can be found in Table \ref{tab:FPRResults}. When examining the FPR, we see that \textit{RedHerring} is able to cause FPRs of 40 to 80 for FGWS, 12 to 69 for WDR, and 19 to 52 for UAPAD. This mirrors the main results as the texts examined should all be labeled as non-attacked, so any labeled as attack are a False Positive. 

The number of successful attacks (those where the classifier predicted correctly, but the detector predicted an attack) can be found in Table \ref{tab:overlapResults}. We see a large number of the attacked examples being successful. For example, in AG News, FGWS detector accuracy is 31.5 which is 685 examples mislabeled, of these 685, 673 are also labeled correctly by the classifier, showing a large overlap. Note, that these values should be taken alongside of the FPR reported above for RedHerring's true effectiveness, as the detectors were not perfectly labeling the 1000 texts before the attack.

These metrics further emphasize the effectiveness of \textit{RedHerring}.

%% file: detectorattack.tex
\subsection{Attack on Detector Only}\label{sect:detectorattack}

\begin{table*}[]
    \centering
    \footnotesize
    \begin{tabular}{c|c||c|c|c|c|c|c}
          & & \multicolumn{6}{c}{Perturbed Accuracy} \\\hline
          
          & Classifier & \multicolumn{3}{c|}{Classifier Acc.} & \multicolumn{3}{|c}{Detection Acc.}\\\hline
         
         \multicolumn{2}{c|}{Target:}   & WDR & FGWS & UAPAD & WDR & FGWS & UAPAD\\\hline

         
         \parbox[t]{2mm}{\multirow{3}{*}{\rotatebox[origin=c]{90}{AG}}} & Albert &  92.0 & 88.8 & 87.7 & 71.8 & 30.0 & 42.0\\
         & RoBERTa & 90.6 & 86.3 & 85.3 & 64.3 & 38.1 & 39.1\\
         & Distilbert &  93.6 & 86.3 & 92.7 & 64.3 & 38.1 & 39.7\\\hline
         \multicolumn{2}{c||}{Average Drops} & 2.4 & 5.0 & 5.9 & 18.9 & 49.8 & 47.0\\\hline\hline

         \parbox[t]{2mm}{\multirow{3}{*}{\rotatebox[origin=c]{90}{RT}}}& Albert & 82.3 & 80.9 & 81.3 & 7.1 & 3.5 & 15.9\\
         & RoBERTa & 80.8 & 78.4 & 80.1 & 15.9 & 9.8 & 32.8 \\
         & BERT & 80.4 & 77.2 & 82.2 & 15.2 & 15.7 & 21.6\\\hline
          \multicolumn{2}{c||}{Average Drops} & 5.1 & 7.4 & 5.1 & 53.3 & 74.2 & 51.6\\\hline\hline
            
        \multicolumn{2}{c||}{Overall Averages} & 3.7 & 6.2 & 5.5 & 36.1 & 62.0 & 49.3\\\hline 
         
    \end{tabular}
    \caption{Results when attack focuses only on detector and does not consider the classifier, as in RedHerring attack.  Target - indicates the target of Detector Attack.  Average drops indicate the average difference between the original and perturbed accuracy values. Overall Averages indicates the average drops in that column. }
    \label{tab:detectorattackResults}
\end{table*}

One goal of \textit{RedHerring} focuses on the classifier maintaining the original label of the text. It may be postulated that this goal is not necessary in this type of attack, instead focusing solely on the detector. This would make the attack more similar to previous adversarial attacks. We measure this, by removing the Goal 2 (in Section \ref{sect:attackgoals}) and testing the attack on Rotten Tomatoes and AG News. The same test texts are used. 

The results for this detector only attack can be found in Table \ref{tab:detectorattackResults}. We find that while this attack is able to drop detector accuracy slightly more than \textit{RedHerring} attack in most cases, it also drops the classifier accuracy at a larger rate. Against UAPAD on AG News, the detector accuracy drops by 38.7 points on average, compared to 35 points by \textit{RedHerring}. However, the classifier accuracy drops by 6 points on average, whereas \textit{RedHerring} causes no drops but rather an increase in classifier accuracy of 2.6 points. This trend is similar across datasets and attack detection algorithms. These results help highlight the importance of the 2nd goal and differentiate \textit{RedHerring} from other, more traditional, adversarial text attacks.

%% file: defense.tex
\section{Initial Defense Against \textit{RedHerring}}\label{sect:defense}

\begin{figure}
    \centering
    \begin{subfigure}[b]{0.95\columnwidth}
        \centering
        \fbox{\includegraphics[width=\textwidth]{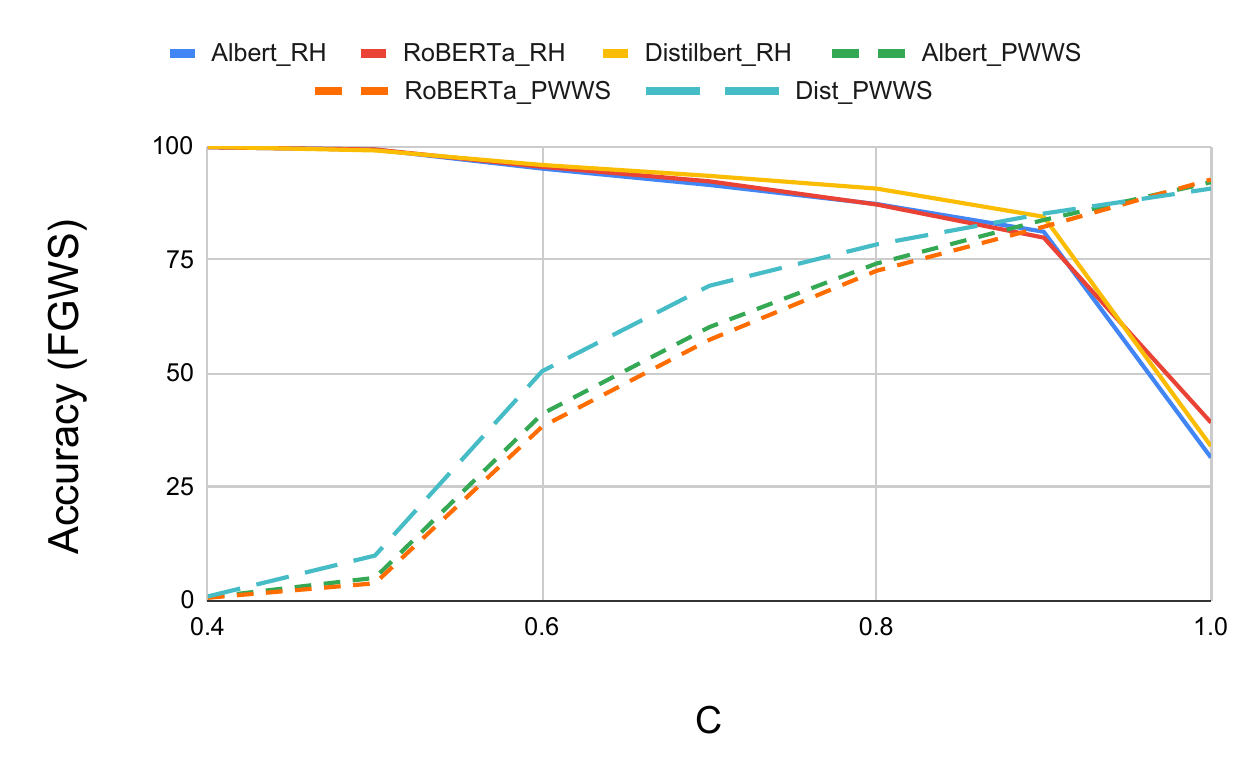}}
    \end{subfigure}
    \caption{Effect of $C$ Values on detection accuracy for the AG News dataset. The solid lines indicate the texts perturbed by \textit{RedHerring}, while the dashed '- -' lines indicate the texts perturbed by PWWS. }
    \label{fig:Ccharts}
\end{figure}

As shown, \textit{RedHerring} poses a serious threat to detectors. To address this, we propose an initial defense which requires no retraining of classifiers or detectors. The defense bases the final detection decision on the confidence of the classifier. Since adversarial attacks aim to reduce the confidence of classifiers to cause the label to flip, classifiers with lower confidence are more likely to have been targeted by an attack. In contrast, \textit{RedHerring} is careful not to drop the classifier's confidence, only the detector's. We propose the ``Confidence Check'' defense found in Appendix \ref{sect:confidencecheck}.

We apply the defense to the WDR and FGWS examples in AG News (Since UAPAD does not return an official detection probability, it is excluded). Furthermore, since we want the algorithm to successfully detect against attacks targeting the classifier, we also test the algorithm on an equal 1000 sample of texts modified by PWWS. These modified texts all successfully tricked the classifier. 

Figure \ref{fig:Ccharts} shows different confidence threshold intervals, $C$, for FGWS (WDR, App. \ref{sect:wdrdefense}). The solid lines are texts modified by \textit{RedHerring} and the other lines (dashed) include those texts modified via PWWS (targeting classifiers). Furthermore, the effect of $C = 1.0$ (for solid lines) are in Table \ref{tab:redherringResults}. 

For detection, we observe a clear tradeoff when exposing the two types of attacks. At $C = 1.0$, max discovery rates are observed for PWWS, but, it causes the \textit{RedHerring} attacked texts to become the least discoverable. As $C$ decreases, we see hinderance in  PWWS accuracy, while \textit{RedHerring} experiences great accuracy increase. Aiming to maximize both detectors, we observe that a $C$ value of $\sim$0.9 to be the best tradeoff.
For FGWS, we ob-\\serve average detection accuracy increases from the low value of 34.9 to the value of 82.6 when examining \textit{RedHerring} texts. For PWWS, we ob-\\serve drops 96.8 to 88.9 on average. For WDR, these increases are similar, but slightly less\footnote{Since the detector was less affected by \textit{RedHerring}. }.

The slight drop in true positives for a large increase in true negatives helps indicate the defense strength, without any retraining of models needed. Though the attack is not fully mitigated, this provides a simple addition to classifier/detector duos to provide some strength against \textit{RedHerring}.

%% file: adversarialtraining.tex
\section{Continuing the Arms Race}
A natural questions is how effective are traditional defenses to \textit{RedHerring}. We explore this by applying adversarial training to WDR. Specifically, we train the detector portion of WDR on the \textit{RedHerring} perturbed texts. We then run \textit{RedHerring} against this new model on a new test set. We test this using 3 classifiers on the Rotten Tomatoes dataset. We use the perturbed test data from the main results for adversarial training and test against 500 additional examples. We compare the adversarially trained model with the original WDR model. The results can be found in Table \ref{tab:adversarialtraining}.  

As can be observed, adversarial training does help mitigate the attack slightly, reducing the effectiveness by 5.4 points for RoBERTa, 8.8 points for Albert, and 8.4 points for T5, but still is vulnerable to \textit{RedHerring}.

\begin{table}[]
    \centering
    \footnotesize
    \begin{tabular}{|c|c|c|c|}
        \hline
          Classifier & Method & Orig. & New \\
          & & Score & Score \\\hline
         RoBERTa & WDR & 77.1 & 11.0 \\
         & AdvWDR & 75.1 & 16.4\\\hline
         Albert & WDR & 55.5 & 2.8 \\
         & AdvWDR & 60.1 & 11.6\\\hline
         T5 & WDR & 75.7 & 4.0 \\
         & AdvWDR & 73.2 & 12.4\\\hline
    \end{tabular}
    \caption{The effectiveness of adversarial training against \textit{RedHerring} on the RT datasets. Orig. Score is the original validation scores (Table \ref{tab:detectionResults}). New Score indicates the model against \textit{RedHerring}.  }
    \label{tab:adversarialtraining}
\end{table}

%% file: conclusion.tex
\section{Conclusion}
\textit{RedHerring} is a novel threat model, which makes attack detectors unreliable. We verify by testing \textit{RedHerring} on 3 attack detectors, across 4 datasets, with 3 - 4 classifiers for each. Overall we see the ability to cause detection drops of 60 points against FGWS on average, 39 points against WDR on average, and 37 points against UAPAD while the classifiers' accuracy is maintained or increased, causing the detectors to become unreliable to a human.


%% file: limitations.tex
\section{Limitations}
Here we note limitations of this research in order to guide future research as well as inform those looking to use this research responsibly:

1. \textbf{Classifiers Explored Limited} Though we examined at least 3 classifiers for each dataset, 3 are built on similar architecture. Though they each achieve high performance in their fields, different models might perform differently or be more sensitive to changes made by \textit{RedHerring} attack.

2. \textbf{More Detection Algorithms and Datasets to Explore} Though we showed effective results against 3 attack detection algorithms, there are more to chooose from. These others may prove more resilient to our proposed attack. Additionally, different datasets could provide difficulty for the attack to be successful against.  

%% file: ethics.tex
\section{Ethical Considerations}
The ethical impacts of studying adversarial attacks must always be weighed, especially when proposing an attack that could lead to unreliability in other models. Especially when malicious users could take and use the research to their own end. However, we believe this research is worth sharing despite the risk for two reasons. First, by exploring this type of attack in this venue, allows others to understand it and build research off to defend against it. Second, we have provided a simple initial defense which requires no retraining of either the classifier or detector. This defense can be easily implemented by those already using the systems. Thus, we believe the learning benefits outweigh the risks.

%% file: relatedwork.tex
\newpage
\onecolumn
\twocolumn
\section{Related Work}\label{sect:relatedwork}
Here we expand on the area of adversarial text attacks and attack detection since \textit{RedHerring} Attack builds off of previous research and targets detection models. 

\noindent \textbf{Adversarial Attacks:} Though adversarial text attacks overlap in goals, there are attributes that divide them. We focus on a few of those attributes relevant to our attack. 

First, the level of the attack is generally divided into 4 categories: 1. Character-level, 2. Word-level, 3. Phrase-level, 4. Sentence-level. Character-level attacks make changes to the text by modifying characters, often with the goal to create unknown tokens. This includes replacing characters with visually similar ones \cite{eger-etal-2019-text}, adding/removing whitespace \cite{Grndahl2018AllYN}, or changing the character order \cite{Li2019Textbugger}. Word-level attacks replace words with less recognized synonyms (from the classifiers' perspective). These attacks have leveraged WordNet \cite{ren-etal-2019-generating}, Word Embeddings \cite{hsieh-etal-2019-robustness}, and Mask Language Models (MLMs) like BERT \cite{garg2020bae, li-etal-2020-bert-attack} to find good substitutions. More recent word-level attacks replace multiple words (or phrases) at once \cite{lei-etal-2022-phrase,deng-etal-2022-valcat}. Sentence-level attacks rewrite texts to cause classifiers to misclassify. Examples include leveraging Machine Translation (MT) \cite{iyyer-etal-2018-adversarial, ribeiro-etal-2018-semantically} to find rewritten texts which a classifier misclassifies or leveraging NLG to change texts at the syntactic level \cite{qi-etal-2021-mind}. Our proposed attack, \textit{RedHerring}, is a word-level attacks. It differs from previous research as it focuses on tricking the detector while keeping the classifier correct, while previous research focuses on tricking the classifier. 

The second aspect in which attacks differ in is level of knowledge available to the attacker. White-box attacks have access to the targeted models' weights and architecture \cite{wang-etal-2022-semattack,sadrizadeh-2022-block}. This allows attacks to quickly find which words in an input text are being used to classify correctly. Black-box attacks have access only to the predicted label and probabilities (or sometimes logits). This limits the attack as it spends more time and queries to find the words which should be replaced \cite{jin2020bert,formento-etal-2023-using}. Our proposed attack assumes the black-box level of knowledge. 

\noindent \textbf{Attack Detection:}
Attack detection algorithms are useful to detect an attack before classification so that the classifier may ignore or pass the text to a human. Most attack detection algorithms examine how the classifier responds to changes in text to help determine if an attack is occuring against that classifier. RS\&V \cite{Wang2021RandomizedSA}, randomly substitutes words with their synonyms at a chosen rate. This occurs $k$ times and the $k$ variations of the text are classified and a majority vote of the label is taken. If the voted label differs from the original classified label, then it is marked as adversarial. WDR \cite{mosca-etal-2022-suspicious} removes words and calculates a score. The scores are given to a classifier to determine if an attack is occuring (described in more detail in Section \ref{sect:attackdetection}). FGWS \cite{mozes-etal-2021-frequency} replaces less frequent words with more frequent synonyms and compares the classification probabilities to determine an attack (described in more detail in Section \ref{sect:attackdetection}). CASN \cite{bao-etal-2023-casn} examines the gradient
of the density data distribution and then calculates the difference between adversarial and normal samples through multiple iterations to detect attacks. UAPAD \cite{gao-etal-2023-universal} utilize universal adversarial perturbations to determine if a text was adversariialy modified or not. In our work we focus on WDR, FGWS, and UAPAD due to wide differences which help create a solid representation of the detection algorithms. Future work will extend attacks to the detection methods leveraging randomness. 

%% file: motivatingexample2.tex
\section{Motivating Example 2}\label{sect:motivatingexample}
 Consider a company that implements spam detection using AI text classifiers. Normally, this spam detector does well on stopping spam. Next, consider a malicious user which wants to bypass the spam detector. They accomplish this by leveraging some adversarial attack algorithm and are able to bypass the spam detector. To combat users like this, the company deploys an adversarial attack detector, which also monitors texts. Now, the attack detector could simply reject all texts it views as attacks, however, this likely could result in too many false positives and users could become frustrated. Instead, when the attack detector labels a text as an attack, it is flagged, and a human-in-the-loop moderator double checks the content.

Now, the malicious user is unable to send spam, even when modified by an adversarial attack algorithm. One way the malicious user could freely send spam again, is to convince the company that the attack detector is faulty. To do this, RedHerring modifies the text instead. When this happens, the classifier labels it as not spam, but the attack detector labels it as an attack, so it is flagged for a human moderator. The human moderator then takes a look at the flagged content, as sees that it is not spam and the classifier classified it correctly as not spam, thus the human moderator approves the content. Now, if this starts to occur very frequently, the attack detector labeling a text as an attack, but the classifier showing no sign of being negatively affected, then the moderator(s) will become less trusting of the detector. This can lead to a) abandoning the detector all together (benefits the malicious user) or b) retraining (or training a new) attack detector (waste of time and money for the company). Even in the second scenario, the attack could continue and cause the company to abandon attack detection altogether as the costs are not justified by the benefits. In the end, it would leave classifiers again more vulnerable to attacks than before and lose security against spam messages.

%% file: uapadweights.tex
\section{UAPAD Weights}\label{sect:uapadweights}
For reproducibility, Table \ref{tab:uapadweights} contains the UAPAD Delta weights for each dataset/classifier combination. 

\begin{table}[]
    \centering
    \footnotesize
    \begin{tabular}{c|c|c}
         & Classifier & Weight \\\hline
         \parbox[t]{2mm}{\multirow{3}{*}{\rotatebox[origin=c]{90}{AG}}} & Albert & 0.2\\
          & RoBERTa & 1.6\\
         & DistilBERT & 0.5\\\hline

         \parbox[t]{2mm}{\multirow{3}{*}{\rotatebox[origin=c]{90}{RT}}} & Albert & 0.2\\
         & RoBERTa & 1.2 \\
         & BERT & 0.7\\\hline

         \parbox[t]{2mm}{\multirow{3}{*}{\rotatebox[origin=c]{90}{IMDB}}} & Albert & 0.3\\
          & RoBERTa & 1.9\\
         & DistilBERT & 0.4\\\hline

         \parbox[t]{2mm}{\multirow{3}{*}{\rotatebox[origin=c]{90}{SST-2}}} & Albert & 0.2\\
         & RoBERTa & 1.3\\
         & DistilBERT & 0.2\\\hline
         
    \end{tabular}
    \caption{UAP Delta Weights used for each dataset/classifer. }
    \label{tab:uapadweights}
\end{table}

%% file: datasetstatistics.tex
\section{Dataset Statistics}
Table \ref{tab:datasetstatistics} contains the distributions of labels for the test examples modified by \textit{RedHerring}. As can be observed, no one class is highly imbalanced against another. 

\begin{table*}[h]
    \centering
    \begin{tabular}{|c|c|c|c|}
        \hline
        \textbf{Rotten Tomatoes} & \textbf{IMDB} & \textbf{SST2} & \textbf{AG News} \\
        \hline
        Pos/Neg & Pos/Neg & Pos/Neg & World/Sports/Business/SciTech \\
        \hline
        533/467 & 503/497 & 428/444 & 268/274/205/253 \\
        \hline
    \end{tabular}
    \caption{Distributions of labels for each dataset.}
    \label{tab:datasetstatistics}
\end{table*}

%% file: originalResults.tex
\section{Original Results}\label{sect:originalresults}
\begin{table*}[]
    \centering
    \footnotesize
    \begin{tabular}{c|c|c|c|c|c|c}
          \multicolumn{7}{c}{Original Accuracy} \\\hline  
        &  & \multicolumn{2}{|c|}{Classifier Acc.} & \multicolumn{3}{|c}{Detection Acc.} \\\hline
         
         & Classifier & \multicolumn{2}{|c|}{ }   & WDR & FGWS & UAPAD \\\hline

         \parbox[t]{2mm}{\multirow{3}{*}{\rotatebox[origin=c]{90}{AG}}} & Albert & \multicolumn{2}{|c|}{94.2} & 86.1 & 81.9 & 90.4\\
         & RoBERTa &\multicolumn{2}{|c|}{94.7} & 85.5 & 80.0 & 85.3\\
         & Distilbert &\multicolumn{2}{|c|}{94.4} & 89.3 & 88.1 & 86.2\\\hline\hline
        
         \parbox[t]{2mm}{\multirow{3}{*}{\rotatebox[origin=c]{90}{IMDB}}}& Albert & \multicolumn{2}{|c|}{93.5} & 80.4 & 89.0 & 73.4\\
         & RoBERTa &\multicolumn{2}{|c|}{95.1} & 87.5 & 92.1 & 87.4\\
         & Distilbert &\multicolumn{2}{|c|}{92.0} & 90.6 & 86.5 & 69.5\\\hline\hline

         \parbox[t]{2mm}{\multirow{3}{*}{\rotatebox[origin=c]{90}{RT}}}& Albert & \multicolumn{2}{|c|}{84.8} & 54.8 & 76.4 & 68.7\\
         & RoBERTa &\multicolumn{2}{|c|}{88.6} & 76.7 & 90.8 & 83.2\\
         & BERT &\multicolumn{2}{|c|}{85.4} & 66.5 & 84.5 & 73.3\\\hline\hline

    
         \parbox[t]{2mm}{\multirow{3}{*}{\rotatebox[origin=c]{90}{SST-2}}}& Albert & \multicolumn{2}{|c|}{92.7} & 57.3 & 83.7  & 57 \\
          & RoBERTa &\multicolumn{2}{|c|}{94.0} & 74.3 & 91.1  & 84.2\\
         & Distilbert &\multicolumn{2}{|c|}{55.7} & 57.0 & 48.7 &60.4\\\hline\hline

    \end{tabular}
    \caption{ Results of the classifiers and detectors on the original test texts which have not been perturbed by RedHerring. RH Target - indicates the target of \textit{RedHerring} Attack}
    \label{tab:originalResults}
\end{table*}

Table \ref{tab:originalResults} contains the original classification and detection results on the datasets with no attack occuring. 

%% file: humanstudyinstructions.tex
\section{Human Study Instructions}\label{sect:humanstudyinstructions}
\begin{table*}[]
    \centering
    \begin{tabular}{p{15.0cm}}
         Instructions  \\\hline
         Has this text been modified to trick the classifier to give the wrong label?\\\\

An attacker may modify text so that a classifier may incorrectly give the wrong label.\\\\
For example, "This movie is great!" may be changed to "This movie is gr8!" so the classifier cannot determine the positive sentiment correctly.\\\\
Another example, "You deserve to die!" might be changed to "You don't deserve to live!" so the classifier cannot determine hatespeech.\\\\

More examples:
"This movie is the worst I've seen!" -> "This movie is the wo rst I've seen!"
"This show is not too bad" -> "This show is n0t too bad!"
"You are a horrible person" -> "You are a heinous human" (Classifier may not be familiar with 'heinous')
\\\\

The attack detector analyzes the text to predict if it has been modified in such a way however the detector is not infallible (it can make mistakes).
It either indicates 'Attack' or 'No attack'.
The attack detector analyzes the text to predict if it has been modified in such a way, however, the detector is not infallible (it can make mistakes).
\\\\
Using the classifier label and detector label, determine if the text has been modified by an attacker.
    \end{tabular}
    \caption{Full Instructions given to human annotators.}
    \label{tab:humaninstructions}
\end{table*}

The full instructions given to the human annotators are given in Table \ref{tab:humaninstructions}. Annotators were asked to label 5 sample texts to make sure they understood the instructions. If any were labeled wrong, they were given explanations before being asked to label the full set of texts. Each annotator labeled 120 texts total and were compensated 20 USD for their time. 

%% file: confidencecheck.tex
\section{Defense Algorithm}\label{sect:confidencecheck}
Algorithm \ref{alg:confidence} contains the algorithm used to defend a detector and classifier against \textit{RedHerring}. 
Formally, let $X$ be an input text. Also, let $g(X)$ be the attack detector and $f(X)$ represent the text classifier. Finally, let $C$ be a confidence threshold defined by the user. If $g(X)$ predicts an attack, ($g(X) == \text{ATTACK}$), then the algorithm checks $f(X)$'s confidence (or probability) given to its predicted class. If $f(X) > C$, then the final prediction is no attack. Otherwise, it is an attack. 
The results for this defense can be found in Section \ref{sect:defense}.

\begin{algorithm}
\caption{Confidence Check}\label{alg:confidence}
\begin{algorithmic}
\Require $X$, $C$
\Ensure $Detect_{pred} \gets \text{\{ATTACK, NOT\}}$
\State $Prob_{pred} \gets \text{f(X)}$
\State $Detect_X \gets g(X)$
\If{$Detect_X == \text{ATTACK}$}
    \If{$Prob_{pred} > C$}
        \State $Detect_{pred} \gets \text{NOT}$
    \Else
        \State $Detect_{pred} \gets \text{ATTACK}$
    \EndIf
\Else
    \State $Detect_{pred} \gets \text{NOT}$
\EndIf
\end{algorithmic}
\end{algorithm}

%% file: defenseresults.tex
\section{Additional Defense Results}\label{sect:wdrdefense}
\begin{figure}
    \centering
    \begin{subfigure}[b]{0.95\columnwidth}
        \centering
        \fbox{\includegraphics[width=\textwidth]{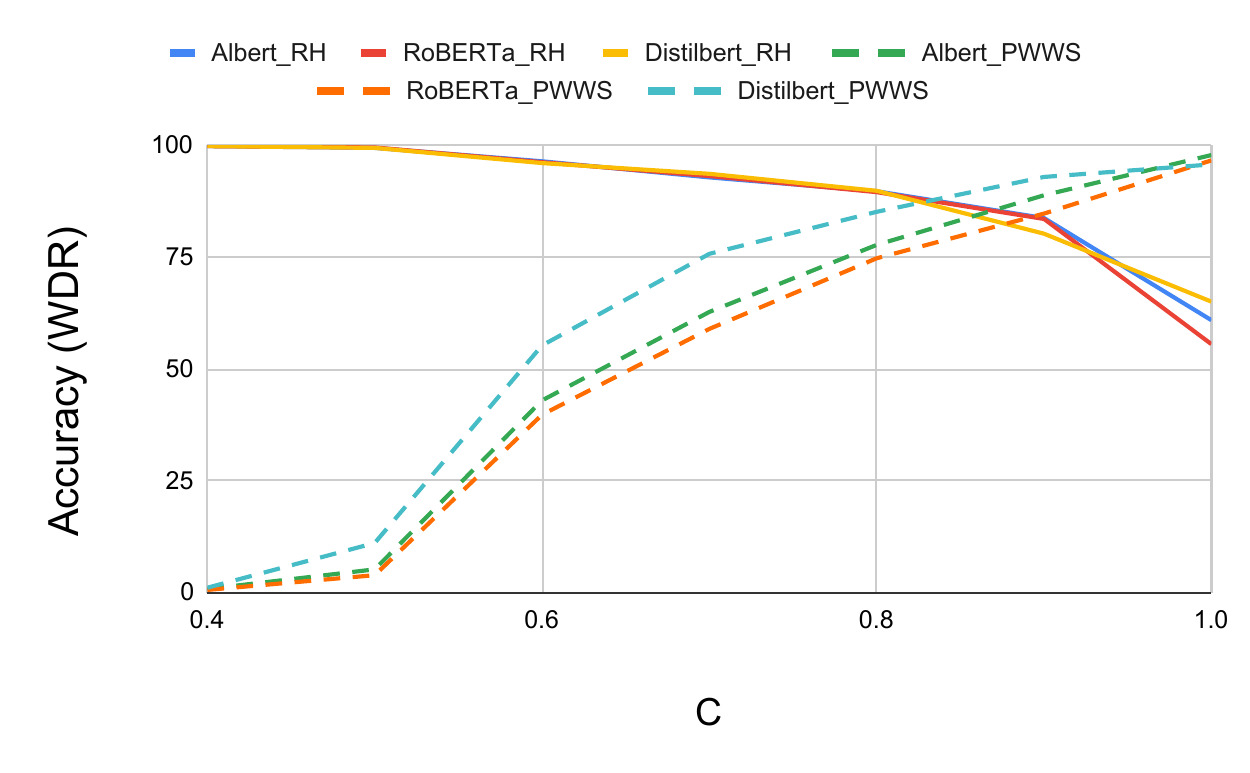}}
    \end{subfigure}
    \caption{Effect of $C$ Values on detection accuracy for the AG News dataset. The solid lines indicate the texts perturbed by \textit{RedHerring}, while the dashed '- -' lines indicate the texts perturbed by an attack targeting the classifiers (PWWS). }
    \label{fig:WDRCcharts}
\end{figure}

The graph for the defense with WDR, can be found in Figure \ref{fig:WDRCcharts}. The defense is described in Section \ref{sect:defense}.

%% file: examples.tex
\section{\textit{RedHerring} Examples}
\begin{table*}[h!]
    \small
    \centering
    \begin{tabular}{p{6.0cm}|p{6.0cm}|p{1.0cm}|p{1.0cm}}
         Original Text & Modified Text  & Class. Conf. & Det. Conf.\\
         \hline
         \hline
          lovingly photographed in the manner of a golden book sprung to life , stuart little 2 manages sweetness largely without stickiness &  lovingly photographed in the manner of a golden book sprung to life, stuart little 2 manages sweetness largely without \textcolor{red}{any} stickiness. & 0.91 & 0.52\\\hline
        throws in enough clever and unexpected twists to make the formula feel fresh.  & throws in enough clever and unexpected twists to make the formula feel \textcolor{red}{more} fresh. & 0.93 & 0.80 \\\hline
         exposing the ways we fool ourselves is one hour photo's real strength. & exposing the ways we fool ourselves \textcolor{red}{-} is one hour photo's real strength. & 0.95 & 0.56\\\hline
         a dreadful day in irish history is given passionate , if somewhat flawed , treatment . & a dreadful day in irish history is given passionate, if somewhat \textcolor{red}{heavily} flawed, treatment. & 0.58  & 0.94\\\hline
        the performances are immaculate , with roussillon providing comic relief  & the performances are \textcolor{red}{generally} immaculate, with roussillon providing comic relief. & 0.83 & 0.66\\
         
    \end{tabular}
    \caption{Examples to illustrate adversarial texts generated by \textit{RedHerring} on the Rotten Tomatoes dataset. Albert is the target classifier and WDR the target detector. Words added are highlighted in \textcolor{red}{red}. Class. Conf. indicates the probability of the classifier for the gold label and Det. Conf. indicates the probability of the detector for the ATTACK label. }
 
    \label{tab:redherringExamples}
\end{table*}

We give examples of text modified by \textit{RedHerring} attack in Table \ref{tab:redherringExamples}. Though this is but a sampling, it helps provide motivation for \textit{RedHerring} meeting its third goal of keeping consistent meaning for humans. We see that for the Rotten Tomatoes examples, \textit{RedHerring} is able to add 1 word in many cases to cause the detector to trigger. Futhermore, these added words flow well in the context of the text. 

We also see that the confidence of the detector varies in the texts as well from 0.52 to 0.94. The classifier confidence for 3 out of 5 examples is greater than 0.9. This also highlights results found in the proposed defense section (Section \ref{sect:defense}). That is, if the defense was applied with a $C$ value of 0.9, then 3/5 would be classified correctly as non-attacks.